\documentclass[10pt,twocolumn,twoside]{IEEEtran}
\usepackage{amsmath,epsfig,setspace,cite,graphicx, multirow}
\usepackage{subfigure}
\usepackage{color}

\begin{document}
%
\title{Block-Matching Convolutional Neural Network for Image Denoising }
%
%

\author{Byeongyong Ahn,
        and~Nam~Ik~Cho,~\IEEEmembership{Senior~Member,~IEEE}
\thanks{B. Ahn and N. I. Cho are with the Dept. of Electrical and Computer Engineering, Seoul National University,
1, Gwanak-ro, Gwanak-Gu, Seoul 151-742, Korea and also
affiliated with INMC
(e-mail: clannad@ispl.snu.ac.kr; nicho@snu.ac.kr).}
}

\markboth{Journal of \LaTeX\ Class Files,~Vol.~6, No.~1, January~2007}%
{Shell \MakeLowercase{\textit{et al.}}: Bare Demo of IEEEtran.cls for Journals}

\maketitle

\begin{abstract}
There are two main streams in up-to-date image denoising
algorithms: non-local self similarity (NSS) prior based methods and 
convolutional neural network (CNN) based methods. 
The NSS based methods are favorable on images with regular and 
repetitive patterns while the CNN based methods perform better on irregular structures. 
In this paper, we propose a block-matching convolutional neural 
network (BMCNN) method that combines NSS prior and CNN. 
Initially, similar local patches in the input image are integrated into a 
3D block. In order to prevent the noise from messing up the block 
matching, we first apply an existing denoising algorithm 
on the noisy image. The denoised image is employed as a pilot signal 
for the block matching, and then denoising function for the block is 
learned by a CNN structure. Experimental results show that the proposed 
BMCNN algorithm achieves state-of-the-art performance. In detail, 
BMCNN can restore both repetitive and irregular structures.
\end{abstract}


\IEEEpeerreviewmaketitle

\section{Introduction}
\label{sec:intro}

With current image capturing technologies, noise is inevitable especially
in low light conditions. Moreover, captured images can be affected by 
additional noises through the compression and transmission 
procedures. Since the noise degrades visual quality, 
compression performance, and also the performance of computer
vision algorithms, the image denoising has been extensively studied
over several decades
\cite{buades2005non,burger2012image,dabov2007image,dong2013nonlocally,
dong2013nonlocal,foi2007pointwise,gu2014weighted,
guleryuz2003weighted,jain2009natural,jia2016adaptive,wang2016adaptive,
sujeong2011,bora}.

In order to estimate a clean image from its noisy 
observation, a number of methods that take account of certain image 
priors have been developed.  Among various image priors,  the NSS is 
considered a remarkable one such that it is employed in most of 
current state-of-the-art methods. The NSS implies that some patterns occur 
repeatedly in an image and the image patches that have similar patterns 
can be located far from each other. The nonlocal means 
filter \cite{buades2005non} is a seminal work
that exploits this NSS prior. The employment of NSS prior has 
boosted the performance of image denoising significantly, and 
many up-to-date denoising algorithms \cite{dabov2007image, mairal2009non, gu2014weighted, nejati2015low, zha2017} can be categorized as NSS 
based methods. Most NSS based methods consist of following steps. 
First, they find similar patches and integrate them into a 3D block. 
Then the block is denoised using some other priors such as low-band prior\cite{dabov2007image}, sparsity prior\cite{nejati2015low, zha2017} 
and low-rank prior\cite{gu2014weighted}. Since the patch similarity can 
be easily disrupted by noise, the NSS based methods are
usually implemented  as two-step or iterative procedures.
 
Although the NSS based methods show high denoising performance, 
they have some limitations. First, since the block denoising stage is designed 
considering a specific prior, it is difficult to satisfy mixed characteristics 
of an image. For example, some methods work very well with the
regular structures (such as stripe pattern) whereas some do not.
Furthermore, since the prior is based on human observation, it can 
hardly be optimal. The NSS based methods also contain some 
parameters that have to be tuned by a user, and it is difficult to find 
the optimal parameters for the overall tasks. Lastly, many NSS based 
methods such as LSSC\cite{mairal2009non}, 
NCSR\cite{dong2013nonlocal} and WNNM\cite{gu2014weighted} 
involve complex optimization problems. These optimizations are
very time-consuming, and also very difficult to be parallelized.

Some researchers, meanwhile, developed discriminative learning 
methods for image denoising, which learn the image prior models 
and corresponding denoising function. Schmidt 
et al.\cite{schmidt2014shrinkage} proposed a cascade of shrinkage 
fields (CSF) that unifies the random field-based model and quadratic 
optimization. Chen et al. \cite{chen2016trainable} proposed a 
trainable nonlinear reaction diffusion (TNRD) model which learns 
parameters for a diffusion model by the gradient descent procedure. 
Although these methods find optimal parameters in a data-driven 
manner, they are limited to the specific prior model. Recently, neural 
network based denoising algorithms 
\cite{jain2009natural, burger2012image, zhang2016beyond} are 
attracting considerable attentions for their performance and
fast processing by GPU. They trained networks 
which take noisy patches as input and estimate noise-free original patch. 
These networks consist of series of convolution operations and 
non-linear activations. Since the neural network denoising algorithms 
are also based on the data-driven framework, they can learn 
at least locally optimal filters for the local regions provided that sufficiently 
large number of training patches from abundant dataset are available. 
It is believed that the networks can also learn the priors which were 
neglected by human observers or difficult to be implemented. However, the
patch-based denoising is basically a local processing and the 
existing neural network methods did not consider the NSS prior. 
Hence, they often show inferior performance than the NSS based methods 
especially in the case of regular and repetitive 
structures \cite{burger2013learning}, which lowers the overall performance.

\begin{figure*}
	\centering
	\begin{tabular}[t]{c}
		\includegraphics[width=14cm]{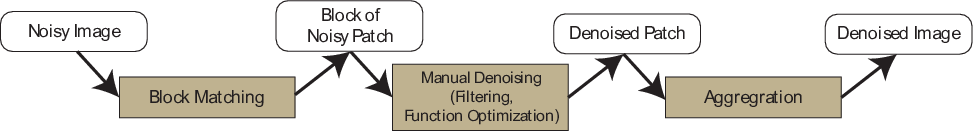}\\
		(a) \\
		\includegraphics[width=14cm]{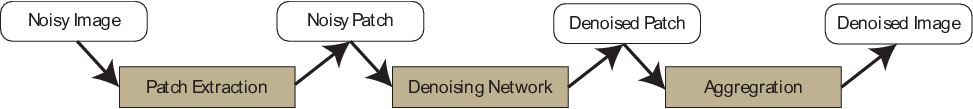}\\
		(b) \\
		\includegraphics[width=14cm]{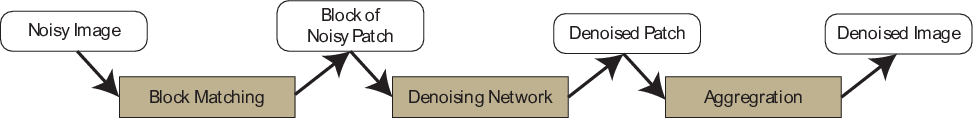}\\
		(c) \\
		\includegraphics[width=6cm]{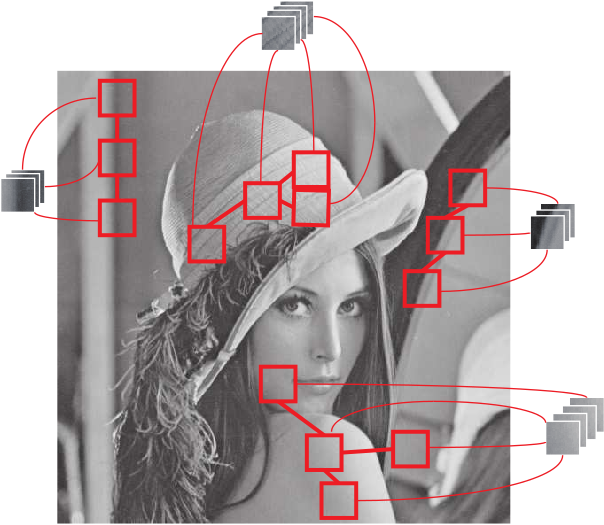}\\
		(d) \\
	\end{tabular}
	\caption{Flow chart of (a) conventional NSS based system, (b)
	NN based system and (c) the proposed BMCNN. 
	(d) Illustration of block-matching step.}
	\label{fig:intro2}
\end{figure*}

In this paper, a combined denoising framework named 
block-matching convolutional neural network (BMCNN) is 
presented. Fig. \ref{fig:intro2} illustrates the difference 
between the existing denoising algorithms and the proposed method. 
As shown in the figure, the proposed method finds similar patches
and stack them as a 3D input like BM3D \cite{dabov2007image},
which is illustreated in Fig. \ref{fig:intro2}-(d). 
By using the set of similar patches as the input, the network 
is able to consider the NSS prior in addition to the local prior 
that the conventional neural networks could train. Compared to 
the conventional NSS based algorithms, the BMCNN is 
a data-driven framework and thus can find more accurate
denoising function for the given input. Finally, it will be explained that 
some of the conventional methods can also be interpreted as a kind of 
BMCNN framework.

The rest of this paper is organized as follows. In Section \ref{sec:Related}, 
researches that are related with the proposed framework are reviewed. 
There are two main topics: image denoising based on NSS, and image 
restoration based on neural network. Section \ref{sec:BMCNN} presents 
the proposed BMCNN network. In Section \ref{sec:Experiments}, 
experimental results and the comparison with the state-of-the-art methods 
are presented. This paper is concluded in Section \ref{sec:Conclusion}.

\section{Related works}
\label{sec:Related}
\subsection{Image Denoising based on Nonlocal Self-similarity}
\label{subsec:Related:NSS}
Most state-of-the-art denoising methods 
\cite{buades2005non, dabov2007image, gu2014weighted, elad2006image,
dong2013nonlocally} employ NSS prior of natural images. The nonlocal 
means filter (NLM)\cite{buades2005non} was the first to
employ this prior to estimate a clean pixel from the relations of
similar non-neighboring blocks.  
Some other algorithms estimate 
the denoised patch rather than estimating each pixel separately. 
For instance, Dabov et al. \cite{dabov2007image} proposed the BM3D 
algorithm that exploits non-local similar patches for denoising a patch. 
The similar patches are found by block matching and they
are stacked to be a 3D block, which is then denoised in the
3D transform domain. Dong et al. 
\cite{dong2013nonlocally, dong2013nonlocal} solved the denoising problem 
by using the sparsity prior of natural images. Since the matrix formed by 
similar patches is of low rank, finding the sparse representation 
of noisy group results in a denoised patch. Nejati et al. \cite{nejati2015low} 
proposed a low-rank regularized collaborative filtering. 
Gu et al.  \cite{gu2014weighted} also considered denoising as a kind of 
low-rank matrix approximation problem which is solved by weighted nuclear 
norm minimization (WNNM). In addition to the low-rank nature, they took 
advantage of the prior knowledge that large singular value of the low-rank 
approximation represents the major components of the image. Specifically, 
the WNNM algorithm adopted the term that prevents large singular values 
from shrinking in addition to the conventional nuclear norm 
minimization (NNM) \cite{recht2010guaranteed}. Recently, 
Zha et al. \cite{zha2017} proposed to use group sparsity residual constraint. 
Their method estimates a group sparse code instead of denoising the 
group directly.

\subsection {Image restoration based on Neural Network}
		
Since Lecun et al. \cite{lecun1998gradient} showed that
their CNN performs very well in digit classification
problem, various CNN structures and related algorithms 
have been developed for diverse computer vision problems 
ranging from low to high-level tasks. Among these, this section 
introduces some neural 
network algorithms for image enhancement problems.
In the early stage of this work, some multilayer perceptrons (MLP) were 
adopted for image processing. Burger et al. \cite{burger2012image} 
showed that a plain MLP can compete the state-of-the-art image 
denoising methods (BM3D) provided that huge training set, deep network and 
numerous neuron are available. Their method was tested on 
several type of noise: Gaussian noise, salt-and-pepper noise, 
compression artifact, etc. Schuler et al. \cite{schuler2013machine} 
trained the same structure to remove the artifacts that occur from 
non-blind image deconvolution. 
		
Meanwhile, many researchers have developed CNN based algorithms. 
Jain et al. \cite{jain2009natural} proposed a CNN for denoising, 
and discussed its relationship with the Markov random field 
(MRF) \cite{geman1986markov}. 
Dong et al. \cite{dong2014learning} proposed a SRCNN, 
which is a convolutional network for image super-resolution. 
Although their network was lightweight, 
it achieved superior performance to the conventional non-CNN approaches. 
They also showed that some 
conventional super-resolution methods such as sparse coding 
\cite{yang2010image} can be considered a special case of deep 
neural network. Their work was continued to the compression 
artifact reduction\cite{dong2015compression}. 
Kim et al. \cite{kim20161,kim20162} proposed two algorithms for image 
super-resolution. In \cite{kim20162}, they presented skip-connection 
from input to output layer. Since the input and output are highly correlated 
in super-resolution problem, the skip-connection was very effective. 
In \cite{kim20161}, they introduced a network with repeated 
convolution layers. Their recursive structure enabled a very deep 
network without huge model and prevented exploding/vanishing
gradients \cite{bengio1994learning}. 

Recently, some techniques such as residual learning 
\cite{he2016deep} and batch normalization \cite{ioffe2015batch} 
have made considerable contributions in developing CNN based image processing 
algorithms. The techniques contribute to stabilizing the convergence of 
the network and improving the performance. For some examples,
Timofte et al. \cite{timofte2014a+} adopted 
residual learning for image super-resolution, and Zhang et al. 
\cite{zhang2016beyond} proposed a deep CNN using both batch 
normalization and residual learning. 

\begin{figure*}	
	\centering
	\includegraphics[width=14cm]{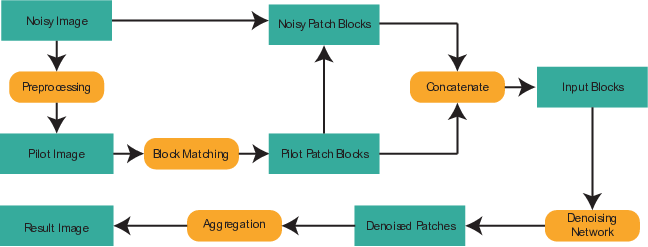}\\
	\caption{The flowchart of proposed BMCNN denoising algorithm.}
	\label{fig:BMCNN_Overview}
\end{figure*}

\section{Block Matching Convolutional Neural Network}
\label{sec:BMCNN}
In this section, we present the BMCNN that esimtates
the original image $X$ from its noisy observation
 $Y = X+V$, where we concentrate on 
additive white Gaussian noise (AWGN) 
that follows $V\sim N(0, \sigma^{2})$.
The overview of our algorithm is illustrated in Fig. \ref{fig:BMCNN_Overview}.
First, we apply an existing denoising method to the noisy image. The 
denoised image is regraded as a pilot signal for block matching. 
That is, we find a group of similar patches from the input 
and pilot images, which is denoised by a CNN. Finally, the denoised 
patches are aggregated to form the output image.

\subsection{Patch Extraction and Block Matching}
\label{subsec:BMCNN_BM}

Many up-to-date denoising methods are 
the patch-based ones, which denoise the image
patch by patch. In the patch-based methods, the
overlapping patch ${\{y_{p}\}}$ of 
size $N_{patch}\times N_{patch}$ are extracted 
from $Y$, centered at the pixel position $p$.
Then, each patch is denoised and merged
together to form an output image. 
In general, this approach yields the best performance when 
all possible overlapping patches are processed, i.e., 
when the patches are extracted with the stride 1.
However, this is obviously computationally demanding and thus 
many previous studies \cite{dabov2007image,burger2012image,gu2014weighted} suggested to use some larger stride that decreases computations while not
much degrading the performance. 

In the conventional NSS based algorithm \cite{dabov2007image},
they first find similar patches to $y_{p}$ based on the 
dissimilarity measure defined as
\begin{equation}
d(y_{p}, y_{q}) = \lVert y_{p} - y_{q} \rVert ^{2}.
\label{eq:BMCNN_blockmatching}
\end{equation}
Specifically, the $k$ patches nearest to 
$y_{p}$ including itself are selected and stacked, which forms a 3D
block $\{Y_{p}\}$ of size $N_{patch}\times N_{patch}\times k$.
Then the block is denoised in the 3D transform domain.
However, it is also shown in  \cite{dabov2007image} that 
the noise affects the block matching performance too much. 
Specifically,  the distance with the noisy observation is a non-central 
chi-squared random variable with the mean
\begin{equation}
E(d(y_{p}, y_{q})) = d(x_{p}, x_{q}) + 2\sigma^{2}N_{patch}^{2}
\end{equation}
and variance
\begin{equation}
V(d(y_{p}, y_{q})) = 8\sigma^{2}N_{patch}^{2}(\sigma^{2}+d(x_{p}, x_{q})).
\end{equation}
where $x_{p}$ and $x_{q}$ are clean image patches that
corresnpond to  $y_{p}$ and $y_{q}$ respectively.
As shown, the variance grows with $O(\sigma^{4})$, and thus the block 
matching results are likely to depend more on the noise 
distribution as the $\sigma$ gets larger. 
This problem is somewhat alleviated by the two-step
approach, i.e., they denoise the 3D block as stated above 
at the first step. Then, at the 
second step, the similar patches are again aggregated by using
the denoised patch as a reference, and the denoised and original
patches are stacked together to be denoised again.

In this respect, we also use a denoised patch to 
find its similar patches from the noisy and denoised
image. Precisely, we adopt an existing 
denoising algorithm as a preprocessing step. The preprocessing step 
finds a denoised image $\hat{X}_{basic}$, which is named 
\textit{pilot image} and used for our patch aggregation as follows:
\begin{itemize}
\item Block-matching is performed on $\hat{X}_{basic}$. Since the 
preprocessing attenuates the noise, the block-matching on 
the pilot image provides more accurate results.
\item The group is formed by stacking both the similar patches 
in the pilot image and the corresponding noisy patches. Since 
some information can be lost by denoising, noisy input patch 
can help reconstructing the details of the image.
\end{itemize}
In this paper, we mainly use DnCNN \cite{zhang2016beyond} to find 
the pilot image because of its promising 
denoising performance and short run-time on GPU. We also use BM3D
as a preprocessing step, which shows almost the same performance
on the average. But the DnCNN and BM3D lead to somewhat different
results for the individual image as will be explained in the experiment
section.

\subsection{Network Structure}

\begin{figure*}	
	\centering
	\includegraphics[width=17cm]{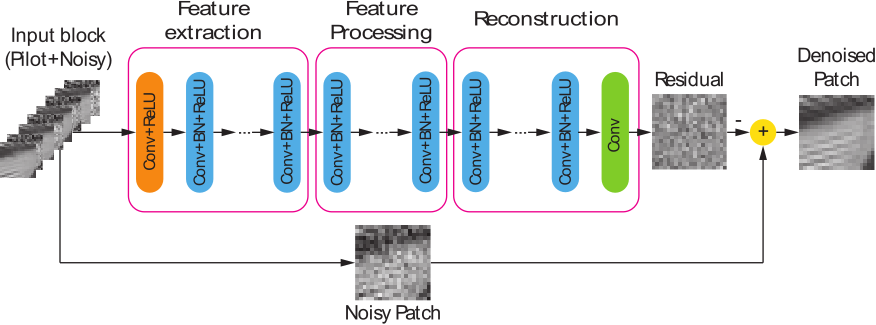}\\
	\caption{The architecture of the denoising network}
	\label{fig:BMCNN_Network}
\end{figure*}

In CNN based methods, designing a network structure is an essential 
step that determines the performance. Simonyan 
et al. \cite{simonyan2014very} pointed out that deep networks consisting 
of small convolutional filters with the size $3\times3$ can achieve favorable 
performance in many computer vision tasks. Based on this principle, 
the DnCNN  \cite{zhang2016beyond} employed 
only $3\times3$ filters, and our network is also
consisted of $3\times3$ filters, with residual learning and batch normalization. 
The architecture of the network is illustrated in Fig. \ref{fig:BMCNN_Network}.

\begin{figure}
	\centering
	\setlength{\tabcolsep}{2.0pt}
	\begin{tabular}[t]{cccccc}
		\includegraphics[width=1.3cm]{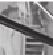}&
		\includegraphics[width=1.3cm]{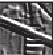}&
		\includegraphics[width=1.3cm]{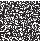}&
		\includegraphics[width=1.3cm]{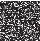}&
		\includegraphics[width=1.3cm]{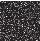}&
		\multirow{9}{*}{\includegraphics[width=1.3cm]{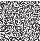}}\\
		\includegraphics[width=1.3cm]{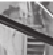}&
		\includegraphics[width=1.3cm]{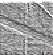}&
		\includegraphics[width=1.3cm]{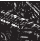}&
		\includegraphics[width=1.3cm]{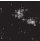}&
		\includegraphics[width=1.3cm]{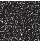}&{}\\
		\includegraphics[width=1.3cm]{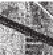}&
		\includegraphics[width=1.3cm]{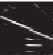}&
		\includegraphics[width=1.3cm]{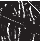}&
		\includegraphics[width=1.3cm]{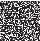}&
		\includegraphics[width=1.3cm]{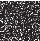}&{}\\
		\includegraphics[width=1.3cm]{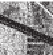}&
		\includegraphics[width=1.3cm]{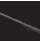}&
		\includegraphics[width=1.3cm]{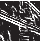}&
		\includegraphics[width=1.3cm]{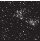}&
		\includegraphics[width=1.3cm]{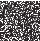}&{}\\
		(a) & (b) & (c) & (d) & (e) & (f)\\
	\end{tabular}
\caption{Feature maps from the denoising network. 
(a) Patches in an input image (b) the output of the first conv layer
(c) the output of the feature extraction stage (at the same time, 
the input to the feature processing stage) (d) the output of the 
feature processing stage (at the same time, the input to the 
reconstruction stage), (e) the input of the last layer, (f) the output 
residual patch.}
\label{fig:BMCNN_Feature}
\end{figure} 

In our algorithm, the depth is set to 17, and the network is composed of 
three types of layers. The first layer generates 
64 low-level feature maps using 64 filters of size $3\times3\times2k$,
for the $k$ patches from the input and another $k$ patches from
the preprocessed image.
Then, the feature maps are processed by a rectified linear unit (ReLU). 
The layers except for the 
first and the last layer (layer 2 $\sim$ 16) contain batch normalization between the convolution 
filters and ReLU operation. The batch normalization for feature maps is proven 
to offer some merits in many previous works
\cite{salimans2016weight,ioffe2015batch,noh2015learning}, such as 
the alleviation of internal covariate shift.  All the convolution operations 
for these layers use 64 filters of size $3\times3\times64$. The last layer 
consists of only a convolution layer. The layer uses a single 
$3\times3\times64$ filter to construct the output from the processed 
feature maps. In this paper, the network adopts 
the residual learning, i.e., $f(Y) = V$ \cite{he2016deep} . Hence, the output 
of the last layer is the estimated noise component of the input and the 
denoised patch is obtained by subtracting the output from the input. 
These layers can also be categorized into three stages as follows.

\subsubsection{Feature Extraction}
At the first stage (layer $1\sim6$), the features of the patches are extracted. 
Figs. \ref{fig:BMCNN_Feature}-(a)$\sim$(c) show the function 
of the stage. The first layer transforms the input patches into the
low-level feature maps including the edges, and then the following layers 
generate gradually higher-level-features. The output of this stage 
contains complicated features and some features about the 
noise components.

\subsubsection{Feature Refinement}

The second stage (layer $7\sim11$) processes the feature maps to construct the 
target feature maps. In existing networks 
\cite{dong2014learning,dong2015compression}, the refinement 
stage filters the noise component out because the main objective is 
to acquire a clean image. On the other hands, the 
target of our algorithm is a noise patch. Hence, the refined 
feature maps are comprised of the noise components as shown in 
Fig. \ref{fig:BMCNN_Feature}-(d). 

\subsubsection{Reconstruction}

The last stage (layer $12\sim17$) makes the residual patch from the
 noise feature maps. 
The stage can be considered an inverse of the feature extraction stage 
in that the layers in the reconstruction stage gradually constructs
lower-level features from high level feature maps as shown in 
Figs. \ref{fig:BMCNN_Feature}-(d)$\sim$(f).
Despite all the layers share the similar form, they contribute 
different operations throughout the network. It gives some intuitions 
in designing an end-to-end network for image processing.

\subsection{Patch aggregation}

In order to obtain the denoised image, it is straightforward to place the 
denoised patches $\hat{x}_{p}$ at the locations of their noisy 
counterparts $y_{p}$. However, as suggested in 
Sec. \ref{subsec:BMCNN_BM}, the step size $N_{step}$ 
is smaller than the patch size $N_{patch}$, which yields an 
overcomplete result consequently. In other words, each pixel is 
estimated in multiple patches. Hence, a patch aggregation 
step that computes the appropriate value of $\tilde{x}(i, j)$ 
from a number of estimates $\hat{x}_{p}(i, j)$ for different $p$ 
is required. The simplest method for the aggregation is simply taking the mean value of the estimates as 
\begin{equation}
\tilde{x}(i, j) = \frac{\sum_{(i, j)\in \hat{x}_{p}}\hat{x}_{p}(i, j)}{\sum_{(i, j)\in \hat{x}_{p}}1}.
\end{equation}
However, in some studies\cite{burger2012image,schuler2013machine}, 
it is shown that weighting the patches $\hat{x}_{p}$ with a simple 
Gaussian window improves the aggregation results. Hence
we also employ the Gaussian weighted aggregation
\begin{equation}
\tilde{x}(i, j) = \frac{\sum_{(i, j)\in \hat{x}_{p}}w_{p}(i, j) \hat{x}_{p}(i, j)}{\sum_{(i, j)\in \hat{x}_{p}}w_{p}(i, j)}
\end{equation}
where the weights are determined as
\begin{equation}
w_{p}(i, j) = \frac{1}{\sqrt{2\pi \sigma_{w}^{2}}} \exp{-\frac{\lvert p-(i, j)\rvert^{2}}{2\sigma_{w}^{2}}}
\end{equation}
where $\sigma_{w}$ is the parameter for weighting. 

\subsection{Connection with Traditional NSS based Denoising}

\begin{figure}
	\centering
	\includegraphics[width=8.5cm]{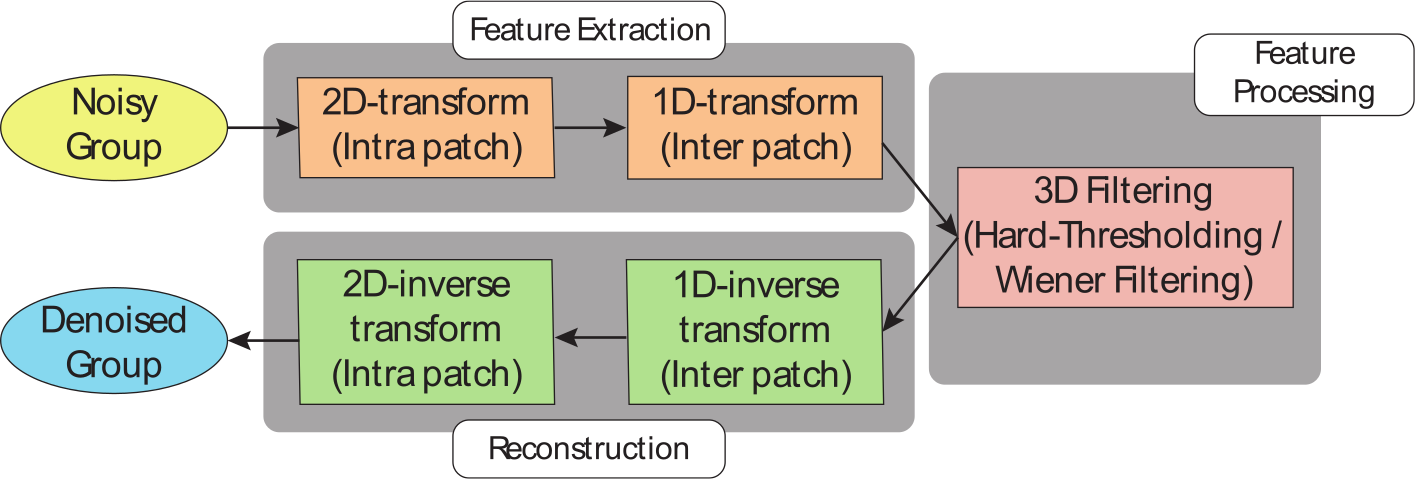}\\
	(a)\\
	\vspace{0.5cm}
	\includegraphics[width=8cm]{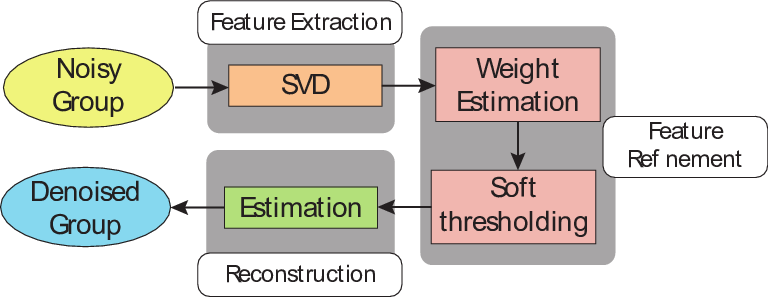}\\
	(b)\\	
	\caption{The group denoising schemes of 
	(a) BM3D and (b) WNNM.}
	\label{fig:BMCNN_NSSmodule}
\end{figure}

In this subsection, we explain that the proposed BMCNN structure can be 
considered a generalization of conventional NSS based algorithms. 
Most existing NSS based denoising algorithms
\cite{dabov2007image,dong2013nonlocal,dong2013nonlocally,gu2014weighted,
jia2016adaptive} share the similar structure: they extract groups of similar 
patches by block matching, denoise the groups separately, and aggregate 
the patches to form the output image. Among the procedures, 
group denoising is the most distinguishing part for individual 
algorithm. In this sense, the analysis 
is focused on the group denoising stage. The group denoising stages of 
state-of-the-art algorithms are demonstrated in 
Fig. \ref{fig:BMCNN_NSSmodule}. 

In BM3D, the input group is first projected to another domain by a 3D
transformation.
Specifically, they perform a 2D tranform to the patches followed by 
a 1D transform into the third dimension. Since the transform
coefficients are often used as features in many image processing algorithms,
the 3D transform corresponds to feature extraction stage. 
Also, as all the transformations employed in the 
BM3D (2D discrete cosine transform (DCT), 2D Bior transform 
and 1D Haar transform) are linear and the parameters are fixed,
the entire 3D transform can be considered a convolution network 
with a single layer of large filter size. In many studies 
on deep neural network \cite{szegedy2015going, simonyan2014very}, 
it has been shown that a convolution layer with large receptive field can 
be replaced by a series of small convolution layers. As a result, the 
feature extraction operator which involves a series of convolution 
can be viewed as a generalization of the 3D transform. After the transform, 
a 3D collaborative filtering is applied to the transformed group. 
Since the filtering attenuates the noise components from each 
coefficient, it corresponds to the feature refinement operation that 
maps noisy features to the denoised ones. The BM3D employs 
hard-thresholding and Wiener filtering for the refinement, where 
both are one-to-one non-linear mapping. In this sense, 
the collaborative filtering behaves as a special case of non-linear 
mapping with $1\times1$ receptive field that can be implemented 
using neural network. 

WNNM\cite{gu2014weighted} transforms the matrix by singular 
value decomposition (SVD). The transformation can be viewed 
as a feature extraction operation that draws some features like 
basis or singluar value from the matrix. The SVD is relatively a complex 
operation compared to the multiplication by a constant matrix such 
as DCT or FFT. In an early work of the neural network 
\cite{bourlard1988auto}, however, it has been explained that an 
optimal solution to an autoencoder is strongly related with the SVD. 
In other word, the SVD decomposition $[U, \Sigma, V] = 
SVD(\{Y_{p}\})$ and the reunion 
$\{\hat{X_{p}}\} = U\hat{\Sigma}V^{T}$ can be successfully 
replaced by the encoder and decoder of an autoencoder network. 
Therefore the proposed feature extraction and patch reconstruction 
operator can be considered a generalization of SVD. 
The singular values are refined by soft-thresholding whose 
weights are determined from the singular value itself. 
Therefore, as a 3D filtering in BM3D, the soft thresholding 
with the weight estimation behaves as a special case of
one-to-one non-linear mapping.

In summary, our BMCNN and many NSS based methods
follow the same process, i.e., block matching followed by
feature extraction and processing in a certain domain by
non-linear mappings. However,
unlike manually deciding the parameters and non-linear 
mapping in the existing methods, 
the proposed BMCNN learns the corresponding procedures 
in a data-driven manner. 
It enables finding the optimal or at least suboptimal 
processing beyond the human design. 
Furthermore, the BMCNN trains an end-to-end mapping that 
consists of all operations rather than considering the operations separately.

\section{Experiments}
\label{sec:Experiments}

\subsection {Training Methodology}
\label{subsec:BMCNN_Training}
The proposed denoising network is implemented using the \textit{Caffe} 
package \cite{jia2014caffe}. Training a network is identical to finding 
an optimal mapping function 
\begin{equation}
\hat{x}_{p} = F(\{W_{i}\}, y_{p})
\label{eq:BMCNN_mapping function}
\end{equation}
where $W_{i}$ is the weight matrix including the bias for the $i$-th layer.
This is achieved by minimizing a cost function 
\begin{equation}
\label{eq:BMCNN_loss}
L(\{W_{i}\}) = \frac{1}{N_{sample}}\sum{d(\hat{x}_{p} , x_{p})} + \lambda r(\{W_{i}\})
\end{equation}
where $N_{sample}$ is the total number of the training samples,
$d(\hat{x}_{p}, x_{p})$ is the distance between the estimated result 
$\hat{x}_{p}$ and its ground truth $x_{p}$,
$r(\{W_{i}\})$ is a regularization term designed to enforce the 
sparseness, and $\lambda$ is the weight for the regularization term. 
Zhao et al\cite{zhao2016loss} proposed several loss functions 
for neural networks, among which we employ the L1 norm for
the distance:
\begin{equation}
\label{eq:BMCNN_individualloss}
d(\hat{x}_{p} , x_{p}) = \sum_{k}{\left| \hat{x}_{p}[k] - x_{p}[k]\right| }
\end{equation}
because of its simplicity for implementation in addition to its 
promising performance for image restoration.
The objective function is minimized using \textit{Adam}, which is
known as an efficient stochastic optimization method \cite{kingma2014adam}. 
In detail, \textit{Adam} solver updates $(W)_{i}$ by the formula
\begin{eqnarray}
(m_{t})_{i}&=&\beta_{1}(m_{t-1})_{i}+(1-\beta_{1})(\bigtriangledown L(W_{t}))_{i},\\
(v_{t})_{i}&=&\beta_{1}(v_{t-1})_{i}+(1-\beta_{1})(\bigtriangledown L(W_{t}))^{2}_{i},\\
(W_{t+1})_{i}&=&(W_{t})_{i}-\alpha \frac{\sqrt{1-(\beta_{2})^{t}}}{1-(\beta_{1})^{t}}\frac{(m_{t})_{i}}{\sqrt{(v_{t})_{i}}+\epsilon}
\end{eqnarray}
where $\beta_{1}$ and $\beta_{2}$ are training parameters, 
$\alpha$ is the learning rate, and $\epsilon$ is a term to avoid 
zero division. In the proposed algorithm, the parameters are 
set as: $\lambda = 0.0002, \alpha = 0.001, \beta_{1} = 0.9, 
\beta_{2}=0.999$ and $\epsilon = 1e-8$. The initial values of 
$(W_{0})_{i}$ are set by Xavier initialization 
\cite{glorot2010understanding}. In \textit{Caffe}, the Xavier initialization 
draws the values from the distribution 
\begin{equation}
(W_{0})_{i}\sim N \left(0, \frac{1}{(N_{in})_{i}} \right)
\end{equation}
where $(N_{in})_{i}$ is the number of neurons feeding into 
the layer. The bias of every convolution layer is initialized to a
constant value 0.2. We train the BMCNN models for three noise 
levels: $\sigma = 15, 25$ and $50$.

\subsection{Training and Test Data}

Recent studies \cite{chen2016trainable,zhang2016beyond} show 
that less than million training samples are sufficient to learn a favorable 
network. Following these works, we use 400 images from the Berkeley 
Segmentation dataset (BSDS) \cite{arbelaez2007berkeley} for the
training. All the images are cropped to the size of $180\times 180$ 
and data augmentation techniques like flip and rotation are applied. 
From all the images, training samples are extracted by the procedure 
in Sec. \ref{subsec:BMCNN_BM}. We set the block size as 
$20\times20\times4$ and the stride as 20. The total number of the training 
samples is 259,200. 

\begin{figure*}
	\centering
	\begin{tabular}[t]{cccccc}
		\includegraphics[width=2.6cm]{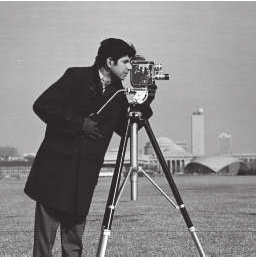}&
		\includegraphics[width=2.6cm]{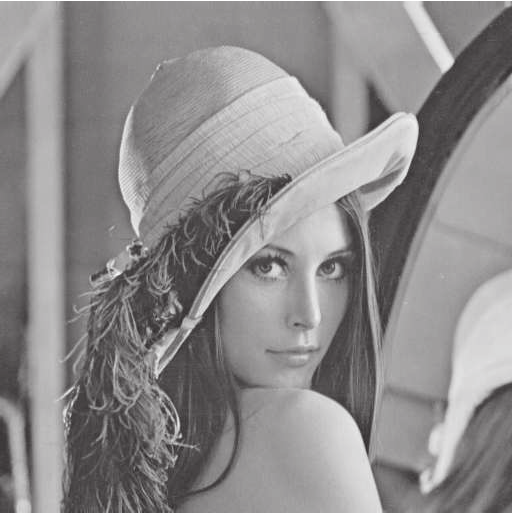}&
		\includegraphics[width=2.6cm]{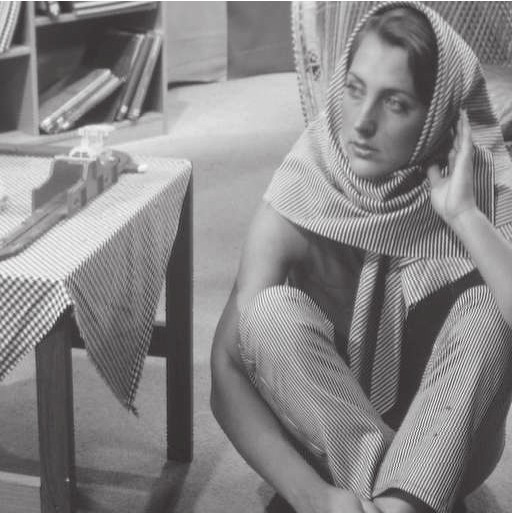}&
		\includegraphics[width=2.6cm]{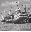}&
		\includegraphics[width=2.6cm]{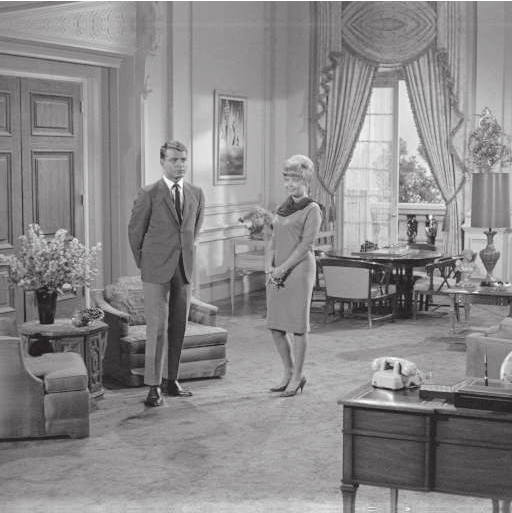}&
		\includegraphics[width=2.6cm]{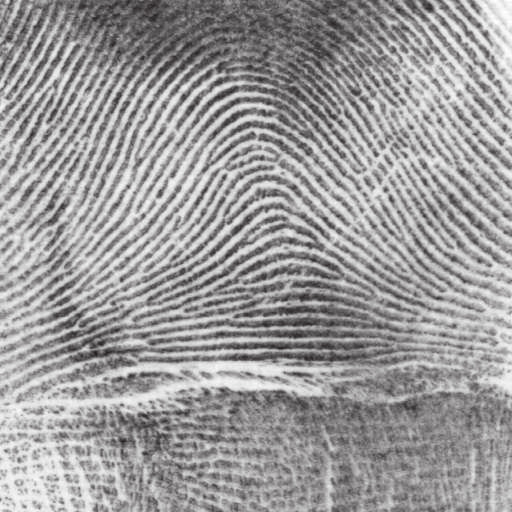}\\
		\includegraphics[width=2.6cm]{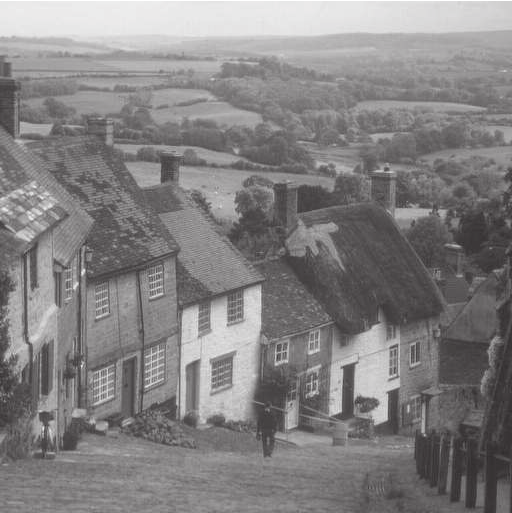}&
		\includegraphics[width=2.6cm]{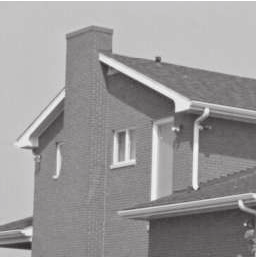}&
		\includegraphics[width=2.6cm]{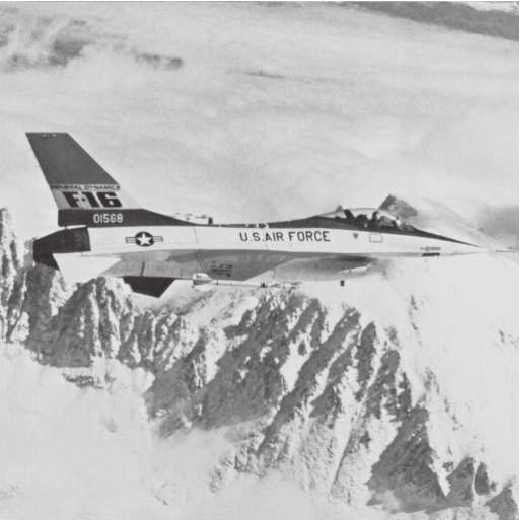}&
		\includegraphics[width=2.6cm]{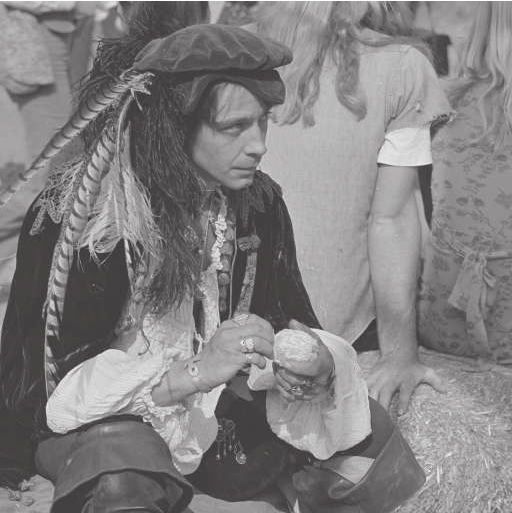}&
		\includegraphics[width=2.6cm]{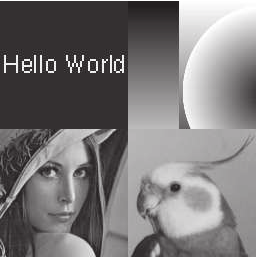}&
		\includegraphics[width=2.6cm]{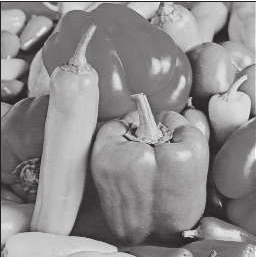}\\
	\end{tabular}
	\caption{The 12 test images used in the experiments}
	\label{fig:BMCNN_Standardimg}
\end{figure*} 

We test our algorithm on standard images that are widely used for 
the test of denoising. Fig. \ref{fig:BMCNN_Standardimg} shows 
the 12 images that constitute the test set. The set contains 4 images 
of size $256\times 256$ (\textit{Cameraman, House, Peppers} and 
\textit{Montage}), and 8 images of size $512\times 512$
(\textit{Lena, Barbara, Boat, Fingerprint, Man, Couple, Hill} 
and \textit{Jetplane}). Note that the test set contains both 
repetitive patterns and irregularly textured images.

\subsection{Comparision with the State-of-the-Art Methods}

In this section, we evaluate the performance of the proposed 
BMCNN and compare it with the state-of-the-art denoising methods, 
including NSS based methods (BM3D \cite{dabov2007image}, 
NCSR\cite{dong2013nonlocally}, and WNNM\cite{gu2014weighted}) 
and training based methods (MLP \cite{burger2012image}, 
TNRD\cite{chen2016trainable}, and DnCNN\cite{zhang2016beyond}).
All the experiments are performed on the same machine - 
Intel 3.4GHz dual core processor, nVidia GTX 780ti GPU and 16GB memory. The testing code for the proposed BMCNN is available at
\textcolor{blue}{http://ispl.snu.ac.kr/clannad/BMCNN}

\subsubsection{Quantitave and Qualitative Evaluation}
\begin{figure*}
	\begin{tabular}[t]{cccc}
		\includegraphics[width=4cm]{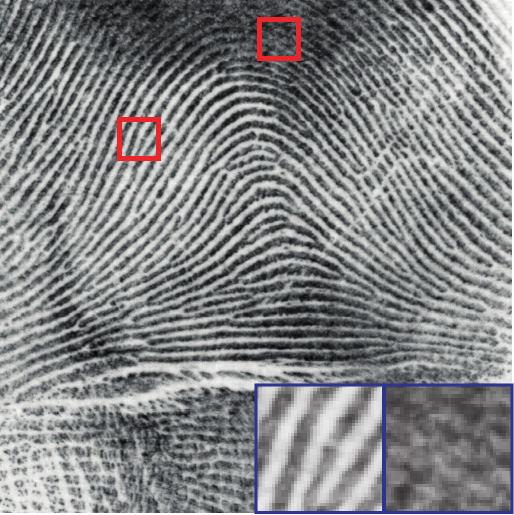}&
		\includegraphics[width=4cm]{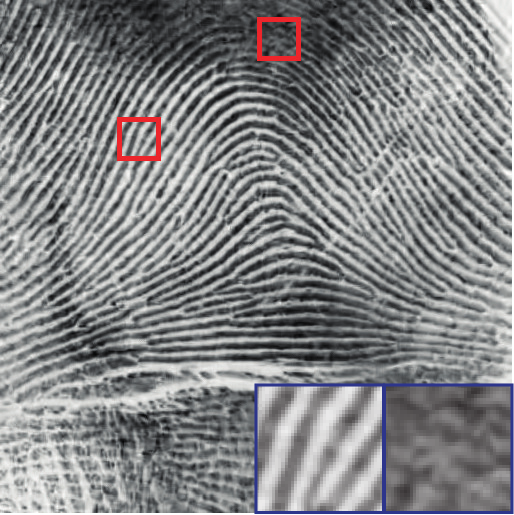}&
		\includegraphics[width=4cm]{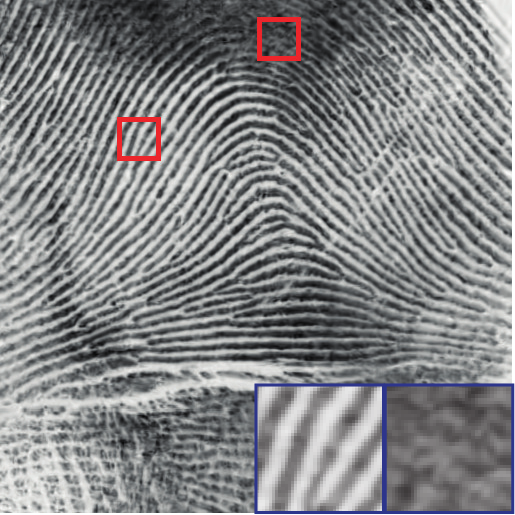}&
		\includegraphics[width=4cm]{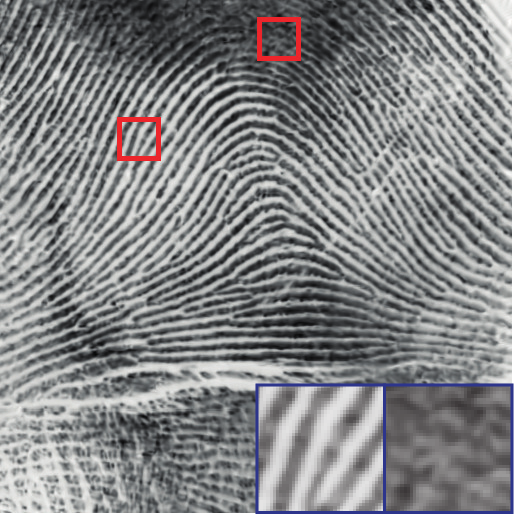}\\
		(a) Original & (b) BM3D & (c) NCSR & (d) WNNM \\
		\includegraphics[width=4cm]{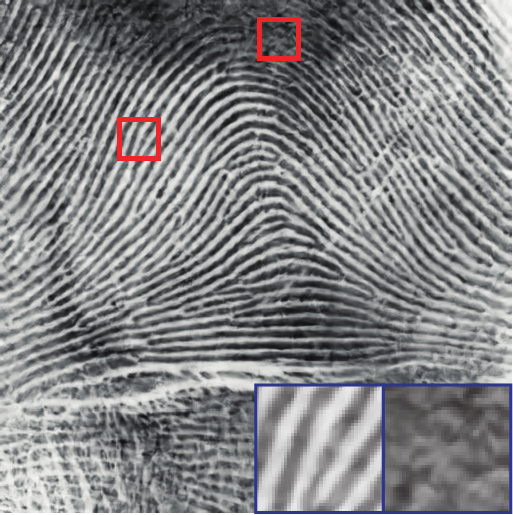}&
		\includegraphics[width=4cm]{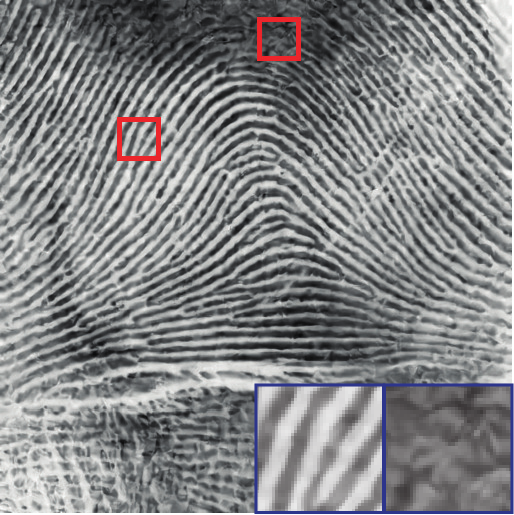}&
		\includegraphics[width=4cm]{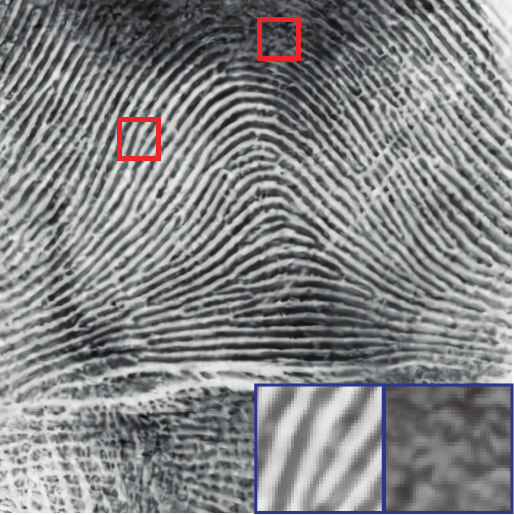}&
		\includegraphics[width=4cm]{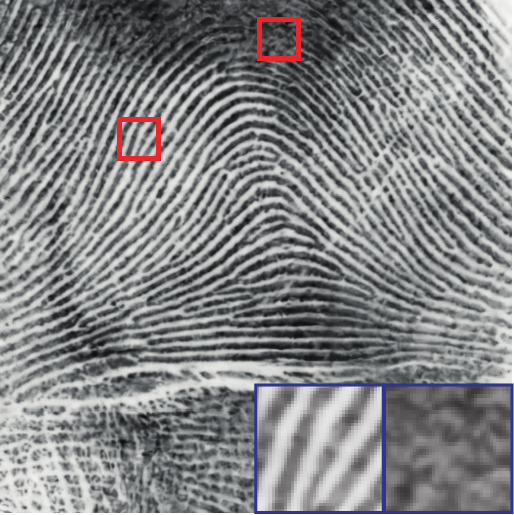}\\
		(e) MLP & (f) TNRD & (g) DnCNN & (h) BMCNN \\	
		\\
	\end{tabular}
	\caption{Denoising result of the \textit{Fingerprint} image with $\sigma$ = 25.}
	\label{fig:BMCNN_Result_1}
\end{figure*}

\begin{figure*}
	\centering
	\begin{tabular}[t]{cccc}
		\includegraphics[width=4cm]{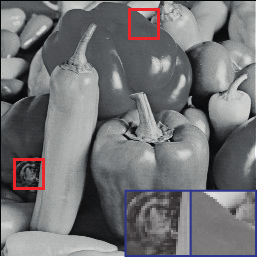}&
		\includegraphics[width=4cm]{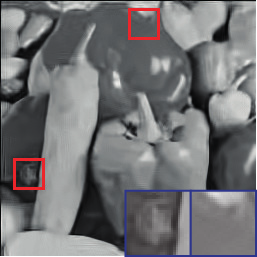}&
		\includegraphics[width=4cm]{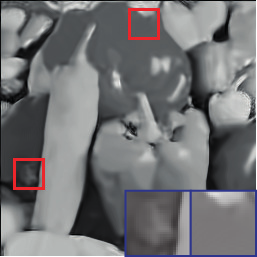}&
		\includegraphics[width=4cm]{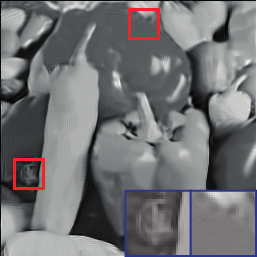}\\
		(a) Original & (b) BM3D & (c) NCSR & (d) WNNM \\
		\includegraphics[width=4cm]{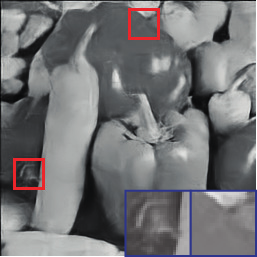}&
		\includegraphics[width=4cm]{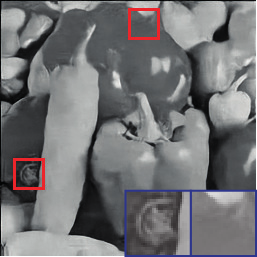}&
		\includegraphics[width=4cm]{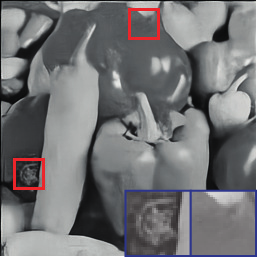}&
		\includegraphics[width=4cm]{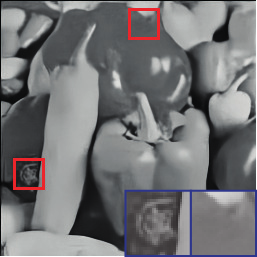}\\
		(e) MLP & (f) TNRD & (g) DnCNN & (h) BMCNN \\	
		\\
	\end{tabular}
	\caption{Denoising result of the \textit{Peppers} image with $\sigma$ = 50.}
	\label{fig:BMCNN_Result_2}
\end{figure*}

The PSNRs of denoised images are 
listed in Table \ref{table:BMCNN_SoA_Performance}. It can be seen
that the proposed BMCNN yields the highest average PSNR for 
every noise level. It shows that the PSNR is improved by 0.1$\sim$0.2dB
compared to DnCNN. Especially, there are large performance gains  
in the case of images with regular and repetitive structure, such 
as \textit{Barbara} and \textit{Fingerprint}, which are the  
images that the NSS based methods perform better than the 
learning based methods. In this sense, it is believed that 
adopting the patch aggregation brings the cons of NSS to
the learning based metod.
Figs.~\ref{fig:BMCNN_Result_1} and \ref{fig:BMCNN_Result_2} 
illustrate the visual results. The NSS based methods tend to blur the
complex parts like a stalk of a fruit and the learning based methods 
often miss details on the repetitive parts such as the stripes of 
fingerprint. In contrast, the BMCNN recovers clear texture in both 
types of regions.

\begin{table*}[!]
	\caption{PSNR of different denoising methods. 
	The best results are highlighted in \textbf{bold}.}
	\vspace{.0cm}
	\label{table:BMCNN_SoA_Performance}
	\centering
	\begin{tabular}{ |c|c|c|c|c|c|c|c|}
		\hline
		Method & BM3D & NCSR & WNNM & MLP & TNRD & DnCNN & BMCNN \\
		\hline
		\multicolumn{8}{c}{ $\sigma=15$}   \\
		\hline
		Cameraman & 31.91 & 32.01 & 32.17 & - & 32.18 & 32.61  & \textbf{32.73}\\
		\hline
		Lena & 34.22 & 34.11 & 34.35 & - & 34.23 &  34.59 & \textbf{34.61} \\
		\hline
		Barbara & 33.07 & 33.03 & \textbf{33.56} & - & 32.11 & 32.60 &  33.08 \\
		\hline
		Boat & 32.12 & 32.04 & 32.25 & - & 32.14 & 32.41 & \textbf{32.42} \\		
		\hline
		Couple & 32.08 & 31.94 & 32.13 & - & 31.89 & 32.40 & \textbf{32.41}\\
		\hline		
		Fingerprint & 30.30 & 30.45 & \textbf{30.56} & - & 30.14 & 30.39 & 30.41\\
		\hline
		Hill & 31.87 & 31.90 & 32.00 & - & 31.89 & \textbf{32.13} & 32.08\\
		\hline
		House & 35.01 & 35.04 & \textbf{35.19} & - & 34.63 & 35.11 & 35.16 \\
		\hline		
		Jetplane & 34.09 & 34.11 & 34.38 & - & 34.28 & \textbf{34.55} & 34.53\\
		\hline
		Man & 31.88 & 31.92 & 32.07 & - & 32.18 & \textbf{32.42} & 32.39\\
		 \hline
		Montage & 35.11 & 34.89 & 35.65 & - & 35.02 & 35.52 & \textbf{35.97} \\
		\hline	
		Peppers & 32.68 & 32.65 & 32.93 & - & 32.96 & 33.21 & \textbf{33.32}\\
		\hline
		\hline
		Average & 32.86 & 32.84  & 33.10 & - & 32.82 & 33.16 & \textbf{33.26}\\
		\hline
		
		\multicolumn{8}{c}{} \\
		[-0.7em]
		\multicolumn{8}{c}{ $\sigma=25$}   \\
		\hline
		Cameraman & 29.44 & 29.47  & 29.64 & 29.59 & 29.69 & 30.11 & \textbf{30.20}\\
		\hline
		Lena & 32.06 & 31.95  & 32.27 & 32.28 & 32.05 & 32.48 & \textbf{32.53}\\
		\hline
		Barbara & 30.64 & 30.57 & \textbf{31.16} & 29.51 & 29.33 & 29.94 & 30.58  \\
		\hline
		Boat & 29.86 & 29.68 & 30.00 & 29.94 & 29.89 & 30.21 & \textbf{30.25}\\		
		\hline
		Couple & 29.69 & 29.46 & 29.78 & 29.72 & 29.69 & 30.10 & \textbf{30.12}\\
		\hline		
		Fingerprint & 27.71 & 27.84 & 27.96 & 27.66 & 27.33 & 27.64 & \textbf{28.01}\\
		\hline
		Hill & 29.82 & 29.68 & 29.96 & 29.83 & 29.77 & 29.99 & \textbf{30.00}\\
		\hline
		House & 32.95 & 32.98 & \textbf{33.33} & 32.66 & 32.64 & 33.23 & 33.32\\
		\hline		
		Jetplane & 31.63 & 31.62 & 31.89 & 31.87 & 31.77 & 32.06 & \textbf{32.17}\\
		\hline
		Man & 29.56 & 29.56 & 29.73 & 29.83 & 29.81 & \textbf{30.06} & 30.06\\
		\hline
		Montage & 32.34 & 31.84 & 32.47 & 32.09 & 32.27 & 32.97 & \textbf{33.47}\\
		\hline	
		Peppers & 30.21 & 29.96 & 30.45 & 30.45 & 30.51 & 30.80 & \textbf{30.93}\\
		\hline
		\hline
		Average & 30.49 & 30.38 & 30.72 & 30.44 & 30.39 & 30.80 & \textbf{30.97}\\
		\hline
		
		\multicolumn{8}{c}{} \\
		[-0.7em]
		\multicolumn{8}{c}{ $\sigma=50$}   \\
		\hline
		Cameraman & 26.18 & 26.15  & 26.47 & 26.37 & 26.56 & 26.99  & \textbf{27.02} \\
		\hline
		Lena & 29.05 & 28.97  & 29.32 & 29.28 & 28.94 & 29.42 & \textbf{29.56}\\
		\hline
		Barbara & 27.08 & 26.93 & \textbf{27.70} & 25.26 & 25.69 & 26.13 &  26.84 \\
		\hline
		Boat & 26.72 & 26.50 & 26.89 & 27.04 & 26.85 & 27.17 & \textbf{27.19} \\		
		\hline
		Couple & 26.42 & 26.19 & 26.59 & 26.68 & 26.48 & 26.88 & \textbf{26.91}\\
		\hline		
		Fingerprint & 24.55 & 24.52 & \textbf{24.79} & 24.21 & 23.70 & 24.14 & 24.65 \\
		\hline
		Hill & 27.05 & 26.87 & 27.12 & \textbf{27.37} & 27.11 & 27.31 & 27.33\\
		\hline
		House & 29.70 & 29.69 & 30.25 & 29.82 & 29.40 & 30.08 & \textbf{30.25}\\
		\hline		
		Jetplane & 28.31 & 28.23 & 28.61 & 28.56 & 28.43 & 28.74 & \textbf{28.88}\\
		\hline
		Man & 26.73 & 26.62 & 26.91 & 27.05 & 26.94 & \textbf{27.18} & 27.17\\
		\hline
		Montage & 27.65 & 27.62 & 27.97 & 28.06 & 28.12 & 29.03 & \textbf{29.50} \\
		\hline	
		Peppers & 26.69 & 26.64 & 26.97 & 26.71 & 27.05 & 27.30 & \textbf{27.45}\\
		\hline
		\hline
		Average & 27.18 & 27.08 & 27.47 & 27.20  & 27.11 & 27.53 & \textbf{27.73}\\
		\hline
	\end{tabular}
\end{table*}

\subsubsection{Running Time}

Table \ref{table:BMCNN_SoA_Time} shows the average run-time 
of the denoising methods for the images of sizes 
$256\times256$ and $512\times512$. For TNRD, DnCNN and 
BMCNN, the times on GPU are computed. As shown, many conventional 
NSS based methods need very long times,
which is mainly due to the complex optimization and/or matrix decomposition. 
On the other hand, since the BM3D consists of simple linear transform 
and non-linear filtering, it is much faster than the NCSR and WNNM. 
Since our BMCNN also consists of convolution and simple ReLU function, 
its computational cost is also less than the WNNM and NCSR. 
The BMCNN is, however, slower than other learning based approaches for three main reasons. First, our algorithm is a two-step approach that 
uses another end-to-end denoising algorithm as a preprocessing step. 
Therefore the computational cost is doubled. Second, the BMCNN contains 
a block matching step, which is difficult to be implement with GPU. 
In our algorithm, the block matching step takes almost half of the run-time. 
Finally, because of the nature of block-matching, the BMCNN is inherently a
patch-based denoising algorithm. In order to prevent the artifacts around 
the boundaries between the denoised patches, the patches are extracted 
with overlapping. Hence, a pixel is processed multiple times and the 
overall run-time increases. Although our algorithm is slower for these 
reasons, our BMCNN is still competitive considering that it is
much faster than the conventional NSS based methods and that
it provides higher PSNR than others.

\begin{table*}[t]
	\caption{Run time (in seconds) of various denoising methods of size $256\times256$ with $\sigma=25$.}
	\vspace{.0cm}
	\label{table:BMCNN_SoA_Time}
	\centering
	\setlength{\tabcolsep}{5pt}
	\begin{tabular}{ |c|c|c|c|c|c|c|c| }
		\hline
		 Methods & BM3D & NCSR & WNNM & MLP & TNRD & DnCNN & BMCNN \\
		\hline
		\hline
		$256\times256$ & 0.87 & 190.3  & 179.9 & 2.238 & 0.038 & 0.053 & 2.135  \\
		\hline
		$512\times512$ & 3.77 & 847.9  & 778.1 & 7.797  & 0.134 & 0.203 & 8.030  \\
		\hline
	\end{tabular}
\end{table*}

\subsection{Effects of Network Formulation}

In this subsection, we modify some settings of the
BMCNN to investigate the relations between the settings and 
performance. All the additional experiments are made 
with $\sigma = 25$.

\begin{table}
	\caption{PSNR results of BMCNN, woBMCNN and their base preprocessing-DnCNN.}
	\vspace{.0cm}
	\label{table:BMCNN_Discussion_BM}
	\centering
	\setlength{\tabcolsep}{5pt}
	\begin{tabular}{ |c|c|c|c| }
		\hline
		 & BMCNN & woBMCNN & DnCNN \\
		\hline
		\hline
		Cameraman & 30.20 & 30.08 & 30.11\\
		\hline
		Lena & 32.53 & 32.45 & 32.48\\
		\hline
		Barbara & 30.58 & 29.91 & 29.94  \\
		\hline
		Boat & 30.25 & 30.17 & 30.21\\		
		\hline
		Couple & 30.12 & 30.09 & 30.10\\
		\hline		
		Fingerprint & 28.01 & 27.61 & 27.64\\
		\hline
		Hill & 30.00 & 29.96 & 29.99\\
		\hline
		House & 33.32 & 33.20 & 33.23\\
		\hline		
		Jetplane & 32.17 & 32.02 &  32.06\\
		\hline
		Man & 30.06 & 30.04 & 30.06 \\
		\hline
		Montage & 33.47 & 33.02 & 32.97\\
		\hline	
		Peppers & 30.93 & 30.81 & 30.80\\
		\hline
		\hline
		Average & 30.97 & 30.78 &  30.80\\
		\hline
	\end{tabular}
\end{table}

\subsubsection{NSS Prior}
In this paper, we take the NSS prior into account by adopting 
block matching, i.e., by using the aggregated similar patches
as the input to the CNN. In order to show the effect of NSS 
prior, we conduct an additional experiment: we train a network 
that estimates a denoised patch using only two patches as the input,
specifically a noisy patch and corresponding pilot patch without further
aggregation. We name the network as woBMCNN, and its PSNR results 
are summarized in Table. \ref{table:BMCNN_Discussion_BM}.
The result validates that the performance gain is owing to the block 
matching rather than two-step denoising. Interestingly, the woBMCNN 
does not perform better than DnCNN, which is used as the preprocessing. 
Actually, the woBMCNN can be interpreted as a deeper network with  
similar network formulation and a skip connection \cite{mao2016image}.
However, since the DnCNN is already a favorable network and the 
performance with the formulation is saturated, the deeper network can 
hardly perform better. On the other hands, the BMCNN encodes 
additional information to the network, which is shown to play an important role.

\begin{table}
	\caption{PSNR results of BMCNN with various patch sizes.}
	\vspace{.0cm}
	\label{table:BMCNN_Discussion_Patch}
	\centering
	\setlength{\tabcolsep}{5pt}
	\begin{tabular}{ |c|c|c|c| }
		\hline
		& $10\times10$ & $20\times20$ & $40\times40$ \\
		\hline
		\hline
		Cameraman & 28.82 & 30.20 & 30.00\\
		\hline
		Lena & 30.42 & 32.53 & 32.39 \\
		\hline
		Barbara & 29.05 & 30.58  & 29.72  \\
		\hline
		Boat & 29.01 & 30.25 & 30.07 \\		
		\hline
		Couple & 28.89 & 30.12 & 29.97 \\
		\hline		
		Fingerprint & 27.24  & 28.01 & 27.51\\
		\hline
		Hill & 28.83 & 30.00 & 29.89\\
		\hline
		House & 30.85 & 33.32 & 33.11\\
		\hline		
		Jetplane & 30.20 & 32.17 &  31.96\\
		\hline
		Man & 28.84 & 30.06 & 29.96 \\
		\hline
		Montage & 30.47 & 33.47 & 32.72 \\
		\hline	
		Peppers & 29.31 & 30.93 & 30.67\\
		\hline
		\hline
		Average & 29.33 & 30.97 &  30.66\\
		\hline
	\end{tabular}
\end{table}

\begin{figure}
	\centering
	\begin{tabular}[t]{ccc}
		\includegraphics[width=2.5cm]{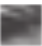}&
		\includegraphics[width=2.5cm]{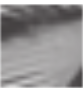}&
		\includegraphics[width=2.5cm]{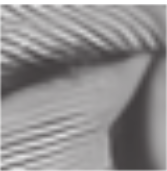}\\
		\includegraphics[width=2.5cm]{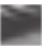}&
		\includegraphics[width=2.5cm]{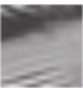}&
		\includegraphics[width=2.5cm]{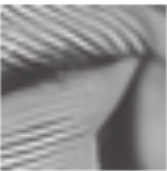}\\
		\includegraphics[width=2.5cm]{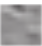}&
		\includegraphics[width=2.5cm]{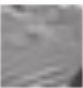}&
		\includegraphics[width=2.5cm]{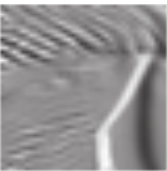}\\
		(a) $10\times10$ & (b) $20\times20$  &(c) $40\times40$ \\
		Error : 0.0515 & Error : 0.0672 & Error : 0.1813 
	\end{tabular}
	\caption{The illustration of block matching results for various patch sizes. 
	The first row shows reference patches, the second row shows the 3rd-similar 
	patches to the references and the third row shows the difference of the 
	first and the second row. The error is defined as the average value of 
	the difference.}
	\label{fig:BMCNN_Disc_Patchsize}
\end{figure} 

\subsubsection{Patch Size}
In many patch-based algorithms, the patch size is an important 
parameter that affects the performance. We train three networks 
with patch size $10\times10$, $20\times20$ (base) and $40\times40$. 
Table \ref{table:BMCNN_Discussion_Patch} shows that $20\times20$
patch works better than other sizes. Moreover, networks of patch 
size $10\times10$ and $40\times40$ 
work even worse than its preprocessing.

Burger et al. \cite{burger2012image} revealed that a larger patch
contains more information, and thus the neural network can learn more 
accurate objective function with larger training patches. On the
contrary, we cannot train the mapping function reasonably with small 
patches. But the large patch degrades the block matching 
performance, because it becomes more difficult to find well
matched patches as the patch size increases. 
Fig. \ref{fig:BMCNN_Disc_Patchsize} shows the block matching result 
and the error for various patch sizes. For a $10\times10$ patch and 
a $20\times20$ patch, the block-matching finds almost the same 
patches and the error is very small. For a $40\times40$ patch, 
on the other hands, decent portion of the patch does not fit well 
and the error becomes so big.  Conventional NSS based algorithms 
including \cite{dabov2007image} and \cite{dong2013nonlocal} 
also prefer small patches whose sizes are around $10\times10$ 
for these reasons. To conclude, $20\times20$ is the proper patch 
size that satisfies both CNN and NSS prior.

\begin{table}
	\caption{PSNR and run time results of BMCNN with various stride.}
	\vspace{.0cm}
	\label{table:BMCNN_Discussion_Stride}
	\centering
	\setlength{\tabcolsep}{5pt}
	\begin{tabular}{ |c|c|c|c|c| }
		\hline
		& 5 & 10 & 15 & 20 \\
		\hline
		\hline
		Cameraman & 30.21 & 30.20& 30.20 &30.18\\
		\hline
		Lena & 32.53 & 32.53 & 32.52 & 32.51\\
		\hline
		Barbara & 30.58 & 30.58 & 30.56 & 30.49 \\
		\hline
		Boat & 30.25 & 30.25 & 30.25 & 30.24\\		
		\hline
		Couple & 30.12 & 30.12 & 30.12 & 30.11\\
		\hline		
		Fingerprint & 28.03 & 28.01 & 28.01 & 27.97\\
		\hline
		Hill & 30.00 & 30.00 & 30.00 & 30.00\\
		\hline
		House  & 33.32  & 33.32 & 33.32 & 33.30\\
		\hline		
		Jetplane  & 32.17 &  32.17 & 32.16 & 32.16\\
		\hline
		Man  & 30.06 & 30.06 & 30.05 & 30.05\\
		\hline
		Montage  & 33.49 & 33.47 & 33.45 & 33.40\\
		\hline	
		Peppers  & 30.94 & 30.93 & 30.93 & 30.92\\
		\hline
		\hline
		Average PSNR & 30.98  &  30.97 & 30.96 & 30.94\\
		\hline
		\hline
		Average time($256\times256$) & 4.271 & 2.151 & 1.765 & 1.603 \\
		\hline
		Average time($512\times512$) & 16.79  & 8.101 & 6.492 & 5.777 \\
		\hline
	\end{tabular}
\end{table}

\subsubsection{Stride}

Since the proposed method is patch-based, its performance  
depends on the stride value to divide the input images into the patches. 
With a small stride, each pixel appears in many patches, which means that 
every pixel is processed multiple times. It definitely increases 
the computational costs but has the possiblity of performance improvement.
We test our algorithm with various stride values and the results are 
summarized in the Table~\ref{table:BMCNN_Discussion_Stride}. 
From the result, we determine that a stride value around the half 
of the patch size shows reasonable performance for both the 
run time and the PSNR.

\begin{table}
	\caption{PSNR of BMCNN results with two different preprocessing.}
	\vspace{.0cm}
	\label{table:BMCNN_Discussion_Pilot}
	\centering
	\setlength{\tabcolsep}{5pt}
	\begin{tabular}{ |c|c|c| }
		\hline
		& BMCNN-DnCNN & BMCNN-BM3D  \\
		\hline
		\hline
		Cameraman & \textbf{30.20} & 30.03 \\
		\hline
		Lena & \textbf{32.53} & 32.49 \\
		\hline
		Barbara & 30.58 & \textbf{31.23}  \\
		\hline
		Boat & \textbf{30.25} & 30.16 \\		
		\hline
		Couple & \textbf{30.12} & 30.02 \\
		\hline		
		Fingerprint & 28.02  & \textbf{28.06} \\
		\hline
		Hill & 30.00 & \textbf{30.05} \\
		\hline
		House & 33.32 & \textbf{33.43} \\
		\hline		
		Jetplane & \textbf{32.17} & 32.14 \\
		\hline
		Man & \textbf{30.06} & 29.94 \\
		\hline
		Montage & \textbf{33.47} & 33.31 \\
		\hline	
		Peppers & \textbf{30.93} & 30.78 \\
		\hline
		\hline
		Average & \textbf{30.97} &  30.97 \\
		\hline
	\end{tabular}
\end{table}

\subsubsection{Pilot Signal}
We also conduct an experiment to see how the different preprocessing 
methods (other than DnCNN in the previous experiments) affect the 
the overall performance.  
For the experiment, we employ BM3D \cite{dabov2007image} due 
to its NSS based nature and reasonable run time. 
Table~\ref{table:BMCNN_Discussion_Pilot} shows the denoising 
performance with different preprocessing methods and two interesting 
characteristics can be found.

\begin{itemize}
\item  The performance on the individual image depends on the preprocessing 
method. BMCNN-BM3D shows better performance on \textit{Barbara, Fingerprint} 
and \textit{House}, where NSS based WNNM performed better than the CNN 
based DnCNN.
\item  However, the overall performance shows negligible difference. 
It implies the overall performance of the denoising network depends on the 
network formulation, not the preprocessing.
\end{itemize}

\section{Conclusion}
\label{sec:Conclusion}
In this paper, we have proposed a framework that combines two dominant 
approaches in up-to-date image denoising algorithms, i.e., the NSS prior 
based methods and CNN based methods. We train a network that estimates 
a noise component from a group of similar patches. Unlike the conventional 
NSS based methods, our denoiser is trained in a data-driven manner to 
learn an optimal mapping function and thus achieves better performance. 
Our BMCNN also shows better performance than the 
existing CNN based method especially in the case of images 
with regular structure, because the BMCNN considers NSS in addition 
to the local characteristics. 

\bibliographystyle{IEEEtran}
\bibliography{IEEEabrv,IEEEexample}

%

%
%
%




\end{document}